\documentclass[10pt, a4paper]{article}

\usepackage{lrec-coling2024}

\usepackage{graphicx}
\usepackage{tabularx}
\usepackage{soul}
\usepackage{xcolor}

\usepackage{bibentry}
\usepackage{latexsym}
\usepackage{xcolor}
\usepackage{adjustbox}
\usepackage{booktabs}
\usepackage{amssymb}
\usepackage{hyperref}
\usepackage{cleveref}
\usepackage{float}

\newcommand{\ignore}[1]{}
\newcommand{\squishlist}{
 \begin{list}{$\bullet$}
  { \setlength{\itemsep}{0pt}
     \setlength{\parsep}{2pt}
     \setlength{\topsep}{2pt}
     \setlength{\partopsep}{0pt}
     \setlength{\leftmargin}{1em}
     \setlength{\labelwidth}{1em}
     \setlength{\labelsep}{0.4em} } }

\newcommand{\squishend}{
  \end{list}  }

\usepackage{xcolor}
 \definecolor{darkblue}{rgb}{0, 0, 0.5}
  \hypersetup{colorlinks=true, citecolor=darkblue, linkcolor=darkblue, urlcolor=darkblue}

\usepackage{xstring}

\usepackage{color}

\title{Emotion Analysis in NLP: \\ Trends, Gaps and Roadmap for Future Directions}

\name{\\\bf {Flor Miriam} {Plaza-del-Arco}$^1$, Alba Curry$^2$,\\ \bf Amanda Cercas Curry$^1$, \bf Dirk Hovy$^1$}
\address{$^1$MilaNLP, Bocconi University, Department of Computing Sciences, Milan, Italy \\
$^2$School of Philosophy, Religion and History of Science, University of Leeds
  \\
  \texttt{\{flor.plaza, amanda.cercas, dirk.hovy\}@unibocconi.it}\\
  \texttt{a.a.cercascurry@leeds.ac.uk}
  }

\abstract{
Emotions are a central aspect of communication. Consequently, emotion analysis (EA) is a rapidly growing field in natural language processing (NLP). However, there is no consensus on scope, direction, or methods.
In this paper, we conduct a thorough review of 154 relevant NLP publications from the last decade. Based on this review, we address four different questions:
(1) How are EA tasks defined in NLP? 
(2) What are the most prominent emotion frameworks and which emotions are modeled?
(3) Is the subjectivity of emotions considered in terms of demographics and cultural factors? and
(4) What are the primary NLP applications for EA?
We take stock of trends in EA and tasks, emotion frameworks used, existing datasets, methods, and applications. We then discuss four lacunae: 
(1) the absence of demographic and cultural aspects does not account for the variation in how emotions are perceived, but instead assumes they are universally experienced in the same manner;
(2) the poor fit of emotion categories from the two main emotion theories to the task;
(3) the lack of standardized EA terminology hinders gap identification, comparison, and future goals; and
(4) the absence of interdisciplinary research isolates EA from insights in other fields. 
Our work will enable more focused research into EA and a more holistic approach to modeling emotions in NLP.
 \\ \newline }
\begin{document}

\maketitleabstract

\section{Introduction}

Emotions perfume our every experience and interaction, playing a key role in human cognition and relationships. 
In recent decades, research on emotion has become popular in numerous fields, including psychology, humanities, and social sciences. In natural language processing (NLP), interest in emotion mostly translates as emotion analysis (EA), whose growth has been most notable since 2018 \cite{strapparava2008learning,mohammad-etal-2018-semeval, klinger-etal-2018-iest, peoples-2020-modeling, MOHAMMAD2021323, wassa-2022-approaches}.

We survey\footnote{Note that our main goal is not a \textit{complete} snapshot of the EA field but rather a snapshot of specific methodological issues within EA in NLP.} over 150 ACL papers (2014-2022)\footnote{For our selection criteria, see Section \ref{sec:survey}.} on EA to address four questions: 
(1) How are EA tasks defined in NLP?
(2) What are the most prominent emotion frameworks, and which emotions are modeled?
(3) Is the subjectivity of emotions considered in terms of demographics and cultural factors?
(4) What are the primary NLP applications for EA?

\begin{figure}[!t]
    \centering
    \begin{adjustbox}{max width=\columnwidth}
    \includegraphics{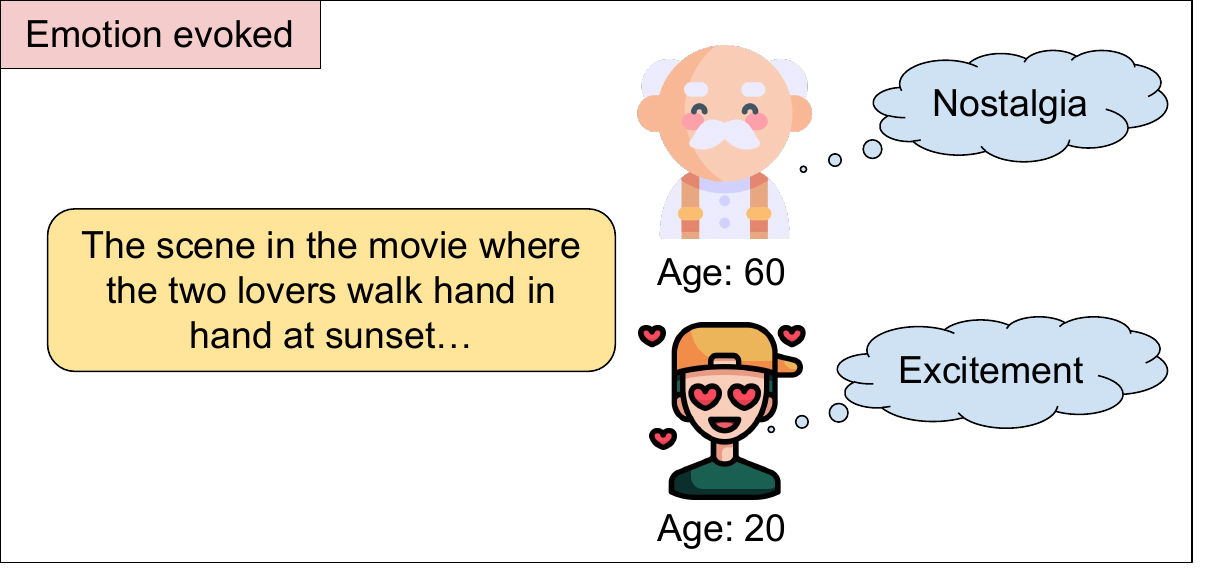}
    \end{adjustbox}
    \caption{Variation in emotion annotation based on demographics. Annotators with distinct demographic profiles, a 60-year-old male and a 20-year-old male, exhibit varying interpretations of the emotions evoked.}
    \label{fig:intro}
\end{figure}

We find a lack of consensus in terms of the scope, direction, and methods in the current studies, which raises three important concerns: 

\squishlist
    \item[1. ] The \emph{absence of demographic and cultural aspects} does not account for the variation in how emotions are presented and perceived but assumes they are universally experienced in the same manner. There is a general lack of engagement with recent work on how emotions are racialized \cite{judd2019sapphire}, gendered \cite{campbell1994being}, etc. In other words, who produces the data matters (see Figure \ref{fig:intro}) and who annotates it.
    \item[2. ] The \emph{poor fit of emotion categories to the task} poses a significant challenge. The commonly used predefined emotion categories may not adequately capture the nuances required for the downstream task. 
    \item[3. ] The \emph{lack of a common systematic nomenclature} in EA obscures gaps, limits comparison, and, therefore, future objectives. For example, our survey shows that `prediction' is synonymous with `classification.' In other areas of NLP, these two terms are used for distinct tasks. 
    \item[4. ] The \emph{dearth of interdisciplinary} research means progress in other fields does not inform EA. A main problematic area is at the level of task design: NLP still predominantly relies on the work of Paul Ekman \cite{EKMANFRIESEN}. Ekman's work addressed the question of  \emph{basic} emotions, not to provide a comprehensive account of emotions nor to define the most `helpful' emotions. Thus, the categories this theory provides are too broad and oversimplified to be useful across tasks and languages \cite{de-bruyne-2023-paradox}. 
\squishend


Emotions are crucial in real-life applications. This paper also explores the diverse range of emotion applications, focusing on dialogue emotion recognition as the most prominent and widely used. We analyze widely used datasets tailored for this purpose, highlighting their key characteristics.


Unlike previous surveys in the field, ours focuses on unexplored questions, revealing critical gaps in the EA field that require community attention. 
We aim to connect these gaps, fostering focused research and nuanced emotion modeling in NLP. We conclude with recommendations and suggestions for future work.





\section{Related Work} 

EA has attracted considerable research interest. Various surveys provide a comprehensive overview of the state of the field \cite{canales-martinez-barco-2014-emotion,10.1145/3057270,10.1145/3123818.3123852,MANTYLA201816,kim2018survey,saxena2020emotion,acheampong2020text}. Recent surveys \cite{murthy2021review,Kusal2022ARO,SINGHTOMAR202394} have primarily focused on the identification of datasets, models, detection techniques, affective computing modalities (visual, vocal, textual, etc.), and applications in the literature. While these surveys acknowledge certain limitations in EA, such as the scarcity of labeled datasets, data imbalance, the dominance of English, and the performance of NLP methods, most of them do not address the significant gaps we discuss.  Specifically, most disregard the significance of demographic factors in their analysis (with a brief mention of the work by \citet{SailunazDRA18}), aggregation methods for collecting annotator perspectives, or the importance of interdisciplinarity.


\citet{mohammad-2022-ethics-sheet} conducted a comprehensive analysis of AI Ethics and Emotion Recognition literature, synthesising 50 ethical considerations relevant to automatic emotion recognition. His work provides valuable ethics pointers that consider some topics we covered here, such as the crucial consideration of demographic factors and the modeling of annotators' perspectives during dataset creation. The gaps we identified demonstrate the inherent connection between ethics and advancements in NLP for EA.

\section{Survey of Emotion Analysis in NLP}
\label{sec:survey}

In this section, we present our survey of papers from the ACL Anthology\footnote{At the time of this study (September 2023), the ACL Anthology contained 88,294 papers, including those from non-ACL events like LREC.}. To ensure a comprehensive selection of studies on EA, we focused on identifying papers whose titles or abstracts include keywords associated with EA NLP tasks in the field. They are `emotion analysis', `emotion prediction', `emotion classification', `emotion detection', `emotion recognition', `emotion polarity classification', and `emotion cause detection.'\footnote{While we acknowledge the existence of other relevant terms in EA, most papers pertinent to our core questions use predominantly these terms.} 
As EA increased in frequency after 2018, we doubled the time horizon to consider papers published in the last ten years. We removed duplicates and papers from shared task participants, resulting in 438 studies. We applied additional criteria to ensure the inclusion of influential and relevant works in our survey. Specifically, we opted for papers with a citation count from Semantic Scholar surpassing the median average (excluding self-citations). However, we adjusted this threshold to at least 1 citation for papers published in the last and current year. Following this selection process, our final analysis set contains 154 papers\footnote{We provide the survey data on GitHub: \url{https://github.com/MilaNLProc/emotion_analysis_survey}.}. 
Among these, 119 are published in main conferences, 31 in workshops, 3 in Findings, and 1 in other venues. For each paper, we collected the following information: 
Emotions used, emotion model type, emotion model, language, resource (dataset, lexicon), data source, presence of multimodality (yes, no), nomenclature used, and applications. If a paper focuses on dataset creation, we also determined whether the annotators' demographics were considered.

\begin{figure}
    \centering
    \begin{adjustbox}{max width=\columnwidth}
    \includegraphics{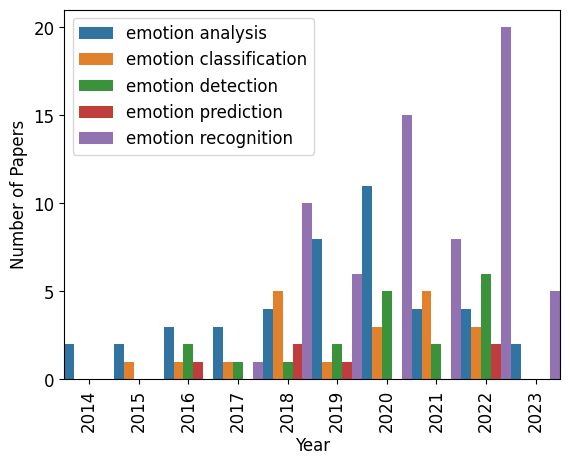}
    \end{adjustbox}
    \caption{Distribution of papers considered in our survey by year and keyword.}
    \label{fig:papers_year_keyword}
\end{figure}

Figure \ref{fig:papers_year_keyword} provides an overview of the 154 papers by year and keyword. There is a clear trend in the number of papers from 2014 to 2022, with 2018 marking a significant increase in the appearance of EA studies. This jump indicates a growing interest in the field during the last five years.

\subsection{Nomenclature}\label{sec:survey_nomenclature}



Figure \ref{fig:papers_year_keyword} also provides an overview of the commonly used terms to describe EA tasks. The term `emotion recognition' stands out as the most frequently used (42.48\%), followed by `emotion analysis' (28.10\%), emotion classification (13.07\%), and `emotion detection' (12.41\%). `Emotion prediction' and `emotion cause detection' are the least used terms.

\begin{figure*}
    \centering
    \includegraphics[width=0.8\linewidth]{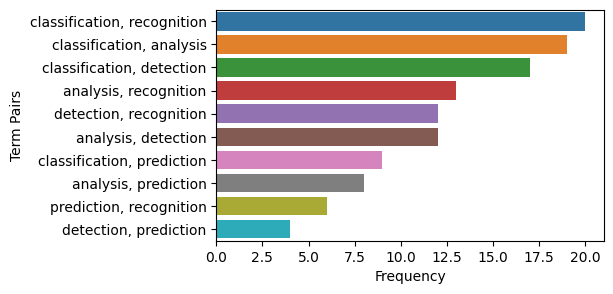}
    \caption{Most common nomenclature pairs for EA tasks used in the literature.}
    \label{fig:term_pairs}
\end{figure*}

However, during our analysis, we note that these terms are not used in isolation. Figure \ref{fig:term_pairs} highlights the most prevalent pairs of terms used across different papers to refer to these tasks. The most frequent pair, `classification-recognition,' is used interchangeably across the papers, followed by `classification-analysis' and `classification-detection.' This variation indicates a lack of standardized terminology for EA tasks. 

Given the lack of uniformity, providing definitions for each term is challenging. We have chosen EA as the broad category that demarcates any emotion-related tasks in NLP. It should be distinguished from sentiment analysis or opinion mining. There is consensus in the literature that sentiment analysis refers to identifying whether someone expresses a positive or negative attitude \cite{10.1145/3057270}. Sentiments are differentiated from emotions by the duration in which they are experienced \cite{6797872}. In contrast, EA is primarily interested in specific emotions such as \texttt{anger} or \texttt{joy}. In the literature reviewed, the task of emotion recognition/detection/classification/prediction refers to the goal of identifying a specific emotion expressed by the author of a particular utterance or set of utterances \cite{mohammad-etal-2018-semeval, chatterjee-etal-2019-semeval}. 
In some cases, it refers to the task of identifying the cause of emotions in a broader context, for example, how speakers affect each other's emotional state in a conversation \cite{liu-etal-2022-dialogueein,zhan-etal-2022-feel}. 
We discuss in Section \ref{sec:terminology} the issues that arise from the lack of consistent terminology. 



% ...

\subsection{Emotion models and frameworks}\label{ssec:emotion_models}



Several conceptual models from psychologists and neuroscientists have emerged to categorize and interpret emotions. Notably, three prominent models stand out: the discrete or categorical model, the dimensional model, and the componential model. They are described in the following:

\begin{itemize}
    \item \textbf{Discrete or categorial model}: Emotions are categorized into distinct classes and categories in this model. Each emotion is treated as independent and separate from others. This model includes well-known frameworks such as Ekman's model \cite{EKMANFRIESEN} consisting of six basic emotions (\texttt{anger}, \texttt{fear}, \texttt{sadness}, \texttt{joy}, \texttt{disgust} and \texttt{surprise}) and Plutchik’s
    model \cite{doi:10.1177/053901882021004003}, which encompasses eight primary emotions (\texttt{anger}, \texttt{anticipation}, \texttt{disgust}, \texttt{fear}, \texttt{joy}, \texttt{sadness}, \texttt{surprise}, and
    \texttt{trust)}.
    \item \textbf{Dimensional model}: Unlike the previous model, dimensional models view emotions as interconnected rather than independent entities. This model acknowledges that emotions can vary in intensity and can be represented along different dimensions. For instance, emotion names are located in vector spaces of affect that consider valence, arousal, and dominance  (VAD) \cite{RUSSELL1977273}.
    
    \item \textbf{Componential model}: The componential model comprised the appraisal theory \cite{scherer1999appraisal, lazarus1991progress}, which emphasizes that emotions are influenced by a person's subjective evaluation or appraisal of a situation or experience. Factors such as \texttt{responsibility}, \texttt{certainty}, \texttt{pleasantness}, \texttt{control}, and \texttt{attention} play a significant role in influencing an individual's emotional responses to a particular situation or experience \cite{smith1985patterns}.
\end{itemize}

\begin{figure}
    \centering
    \begin{adjustbox}{max width=\columnwidth}
    \includegraphics{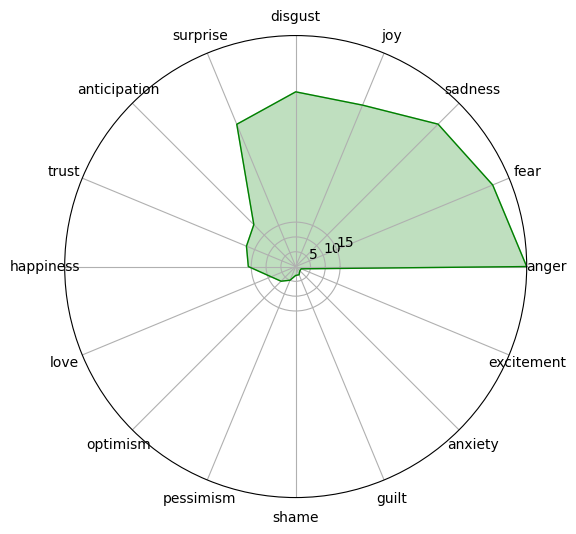}
    \end{adjustbox}
    \caption{Most common emotion categories used in the EA literature.}
    \label{fig:emotion_frequencies}
\end{figure}

Our analysis reveals a clear prevalence of discrete conceptual models in the literature, indicating a strong interest in classifying emotions into distinct categories and treating each emotion as an independent entity -- in line with standard supervised learning objectives. However, a few papers use dimensional models, and an even smaller proportion explore componential models. 

Research papers in discrete or categorical emotion models commonly rely upon different theories. The most influential theory is the Ekman model \cite{EKMANFRIESEN}, followed by the Plutchik model \cite{doi:10.1177/053901882021004003}. A few other studies considered the Turne list \cite{turner2000origins} and the Ortony, Clore, and Collins model \cite{ortony1988}. 
With dimensional models, some studies used a mapping approach to associate emotion categories with dimensions. For instance, \citet{bagher-zadeh-etal-2018-multimodal} conducted data annotating mapping the Ekman emotions to a Likert scale in which values are assigned from 0 to 3 to represent the presence of emotions. \citet{park-etal-2021-dimensional} introduced a framework that leverages VAD scores obtained from an emotion lexicon to acquire VAD scores from sentences labeled with categorical emotions. More recently, componential models have gained attention and been introduced in the field. \citet{hofmann-etal-2020-appraisal} established an annotation task where event descriptions are linked to seven appraisal dimensions \cite{smith1985patterns}. 
\citet{stranisci-etal-2022-appreddit} relied on Roseman's model of appraisal \cite{roseman1991appraisal} to annotate a corpus of social media posts. More recently, \cite{troiano-etal-2023-dimensional} built a corpus where people described emotion-triggering events and their appraisals. Subsequently, readers were asked to reconstruct emotions and appraisals from the text. They provide a thorough appraisal theory overview for text-based emotion analysis suitability assessment. Additionally, it shows that appraisals offer an alternative computational approach to EA and enhance emotion categorization in text using joint models.


Figure \ref{fig:emotion_frequencies} provides an overview of the most frequent emotion categories in the literature. Due to the widespread use of Ekman's model \cite{EKMANFRIESEN}, the prevalent emotions observed are \texttt{anger}, \texttt{fear}, \texttt{sadness}, \texttt{joy}, \texttt{disgust}, and \texttt{surprise}. Additionally, Plutchik's model, which is the second most used, highlights the frequency of \texttt{trust} and \texttt{anticipation}. However, a range of other emotions receives comparatively less attention. These include \texttt{excitement}, \texttt{anxiety}, \texttt{optimism}, \texttt{pessimism}, \texttt{guilt}, \texttt{shame}, and \texttt{love}.



\begin{table*}[t]
  \centering
  \small
  \begin{tabular}{p{4.5cm}|p{4cm}|l|l|p{1.5cm}}
    \toprule
    \textbf{Dataset} & \textbf{Annotation} & \textbf{Lang.} &  \textbf{Source} & \textbf{Size} \\
    \midrule
     IdeDialog \cite{ide-kawahara-2022-building} & P & JA & Twitter dialogues & 13,806 \\
    \midrule
    M3ED \cite{zhao-etal-2022-m3ed} & E + neutral & CN & TV dialogues & 24,449 \\
    \midrule
    MuSE \cite{jaiswal-etal-2020-muse} & sad, anger, contentment, amusement, neutral & EN  & Recordings & 28 speakers \\
    \midrule
    MELD \cite{Zahiri2017EmotionDO,poria-etal-2019-meld} & E + neutral & EN & Movie dialogues & 13,000 \\
    \midrule
    DailyDialog \cite{li-etal-2017-dailydialog}  & E + other & EN  & Movie dialogues & 13,118 \\
    \midrule
    EmoryNLP \cite{Zahiri2017EmotionDO} & E + neutral & EN &  TV show dialogues & 12,606 \\
     \midrule
    IEMOCAP \cite{busso2008iemocap} & happiness, anger, sadness, frustration, neutral & EN & Dyadic interactions & 10,039  \\
    \midrule
    \midrule
    StudEmo \cite{ngo-etal-2022-studemo} &  28 emotion categories & EN & Customer reviews &  5,182 \\
    \midrule
    RED v2 \cite{ciobotaru-etal-2022-red} & P - disgust, anticipation & RO &Tweets  & 5,449 \\
    \midrule
    APPReddit \cite{stranisci-etal-2022-appreddit} & appraisals \cite{roseman1991appraisal} & EN &  Reddit & 1,091 \\
    \midrule
    Universal Joy \cite{lamprinidis-etal-2021-universal} & anger, anticipation, fear, joy, sadness & 18 & Facebook posts & 530k \\
    \midrule
    WRIME \cite{kajiwara-etal-2021-wrime} & P, emotion intensity & JA &  SNS posts & 17,000 \\
    \midrule
    EmoEvent \cite{plaza-del-arco-emoevent} & E + neutral & EN-ES &  Tweets & 8,409 \\
    \midrule
    GoodNewsEveryone \cite{bostan-etal-2020-goodnewseveryone} & 15 emotion categories & EN &  News Headlines & 5,000 \\
    \midrule 
    PO-EMO \cite{haider-etal-2020-po} & joy, sadness, uneasiness, vitality, suspense, sublime, humor, annoyance, nostalgia & EN-DE &  Poems & 64 (EN), 18 (DE) \\
    \midrule 
    enISEAR \cite{hofmann-etal-2020-appraisal} & appraisals \cite{smith1985patterns} & EN &  Self-reports & 1,001 \\
     \bottomrule
  \end{tabular}
  \caption{EA datasets. [E] Ekman, [P] Plutchik. The first part of the table contains multimodal datasets used for emotion recognition in conversation. The second part contains datasets used in EA and recent ones. Lang.: Language}
  \label{tab:datasets}
\end{table*}

\subsection{Data}\label{ssec:data}



In this section, we provide an overview of the datasets that have been extensively used in various EA studies and more recently developed datasets. In our analysis, we identified over 50 papers that created a dataset. Table \ref{tab:datasets} gives a summary list of some of these datasets, showcasing their key characteristics: annotation scheme, language, multimodality, source, and size.

Most of the datasets found in the EA studies are focused on emotion recognition in conversation, including IdeDialog \cite{ide-kawahara-2022-building}, M3ED \cite{zhao-etal-2022-m3ed},  MuSE \cite{jaiswal-etal-2020-muse}, MELD \cite{Zahiri2017EmotionDO,poria-etal-2019-meld}, DailyDialog \cite{li-etal-2017-dailydialog}, EmoryNLP \cite{Zahiri2017EmotionDO}, and IEMOCAP \cite{busso2008iemocap}. The annotation scheme used in most of these datasets follows the discrete or categorical model, specifically the Ekman model theory. However, IdeDialog uses the Plutchik model, which encompasses different emotion categories. Additionally, IEMOCAP and MuSE use their own unique annotation scheme. In terms of language, most of these datasets feature primarily English dialogues, except for M3ED, which incorporates Chinese dialogues, and IdeDialog, which includes Japanese dialogues. These datasets are multimodal, including text, audio, and visual information. Moreover, the dialogues in these datasets are sourced from diverse domains, including movies, dyadic interactions, TV shows, Twitter, and recordings. 

In the second part of the table, we summarize some of the recent EA datasets created over the past few years. These datasets include StudEmo \cite{ngo-etal-2022-studemo}, RED v2 \cite{ciobotaru-etal-2022-red}, APPReddit \cite{stranisci-etal-2022-appreddit}, Universal Joy \cite{lamprinidis-etal-2021-universal}, WRIME \cite{kajiwara-etal-2021-wrime}, EmoEvent \cite{plaza-del-arco-emoevent}, GoodNewsEveryone \cite{bostan-etal-2020-goodnewseveryone}, PO-EMO \cite{haider-etal-2020-po}, and enISEAR \cite{hofmann-etal-2020-appraisal}. The annotation schemes of these datasets offer a wider range of emotion frameworks and emotions compared to the previously described for emotion recognition in conversation. While most adhere to the discrete emotion model, enISEAR and APPReddit incorporate the componential model with appraisals. Although English is the most predominant language, multilingual datasets like EmoEvent and Universal Joy encompass multiple languages. RED includes Romanian, and WRIME includes Japanese, further diversifying the linguistic coverage of emotion recognition studies. None of these datasets present a multimodal aspect. The instances are sourced from diverse domains, including social media (Twitter, Facebook), news headlines, self-reports, poems, customer reviews, and community networks (Reddit). Among these domains, social media emerges as the most predominant data source.

Despite the highly subjective nature of emotions, only a few works \cite{martin-etal-2022-swahbert,zhan-etal-2022-feel,ghosh-etal-2022-comma} mentioned the gathering of annotator demographic information, yet most of them do not use this information for their methods.

\subsection{Applications}

Studying patterns of human emotions and understanding people's feelings are essential in many real-life applications \cite{picard2000affective}. In our analysis, among the most used applications, we found that emotion recognition in conversation has gained significant popularity in the field by analyzing the emotions expressed during interactions \cite{hazarika-etal-2018-icon, zhong-etal-2019-knowledge, ghosal-etal-2020-cosmic, hu-etal-2021-dialoguecrn, lee-lee-2022-compm}. Another noteworthy application within the realm of EA is its application in the analysis of narrative texts \cite{kim-klinger-2018-feels,liu-etal-2019-dens,haider-etal-2020-po,zad-finlayson-2020-systematic, cortal-etal-2023-emotion}. Additionally, EA has found a valuable niche in the healthcare sector, particularly in addressing mental health concerns \cite{khanpour-caragea-2018-fine}. With the advent of social media, detecting and monitoring emotional states associated with mental well-being has become possible. This capability has proven invaluable in identifying individuals at risk of self-harm, anxiety, or stress, as detailed by \citet{turcan-etal-2021-emotion}. EA also proves beneficial during critical events like the COVID-19 pandemic \cite{ng-etal-2020-miss,sosea-etal-2022-emotion,sampath-etal-2022-findings}. 
Beyond these applications, EA extends its impact to various other domains including email customer care \cite{gupta-etal-2010-emotion}, span prediction \cite{alhuzali-ananiadou-2021-spanemo}, fake news \cite{melleng-etal-2019-sentiment}, moral content detection \cite{asprino-etal-2022-uncovering}, sarcasm detection \cite{chauhan-etal-2020-one,chauhan-etal-2020-sentiment}, time series analysis of emotional loading in central bank statements \cite{buechel-etal-2019-time}, and financial forecasting \cite{seroyizhko-etal-2022-sentiment}.

\section{Shortcomings of EA in NLP}
In this section, we outline the main areas where there is room for improvement in EA research. 

\subsection{Lack of Diversity in Available Datasets} 
We find significant overlap between datasets: most are in English, sourced from movies or TV shows, and annotated using Ekman's emotions categorically. The lack of language diversity in NLP has been discussed at length \cite{martin2022swahbert,sonu2022identifying}, and it is not specific to EA, but EA lacks diversity in other aspects: 

\paragraph{Diversity in emotion labels.} 
\Cref{ssec:data} shows that most datasets are labeled according to Ekman's basic emotions, or Plutchik to a lesser extent. While these emotions are widely used in the analyzed studies, they do not generalize to all tasks or domains \cite{de-bruyne-2023-paradox}. Human emotions are nuanced because this nuance is required for understanding the world. Yet current datasets predominantly focus on a few coarse-grained emotions. For example, feeling guilty or lonely both fall under sadness but convey very different meanings. Furthermore, the most commonly annotated emotions are negative, particularly \texttt{anger}, \texttt{fear}, and \texttt{sadness}.  This emphasis on negative emotions has far-reaching consequences for EA and downstream applications.
Detecting more fine-grained negative emotions might be helpful in a mental disorder detection setting, but positive emotions may be more relevant for tutoring systems as they aid learning. A notable example demonstrating the importance of aligning emotion categories with the specific task at hand is the work of \cite{haider-etal-2020-po}, where aesthetic emotions play a pivotal role in analyzing emotions within poems. This approach encompasses emotions such as \texttt{joy}, \texttt{sadness}, \texttt{uneasiness}, \texttt{vitality}, \texttt{suspense}, \texttt{sublime}, \texttt{humor}, \texttt{annoyance}, and \texttt{nostalgia}.. It is also noteworthy that most of the papers in the survey did not specify any concrete application of EA -- lacking guidance in selecting the labels that will aid in a given task. 

\paragraph{Diversity in annotation schemes.} 

The preference for Ekman's or Plutchik's models significantly limits the adaptability of EA techniques to different contexts and tasks \cite{plaza-del-arco-natural}. Each NLP task may require a nuanced understanding of emotions that these models might not fully capture. This limitation impedes the development of more versatile and context-specific emotion analysis approaches.
Furthermore, this preference for specific models neglects the rich landscape of existing psychological theories related to emotions. Emotion is a complex and multifaceted phenomenon, and various psychological theories offer different insights into how emotions are generated, expressed, and perceived \cite{hofmann-etal-2020-appraisal}. Overlooking these diverse theories means missing out on knowledge, valuable perspectives, and dimensions of emotion. More recent studies follow alternative theories of cognitive appraisal of events (see \Cref{ssec:emotion_models}) and show their potential for emotion classification when encoded in categorical models. \citealt{ohman2020emotion} also point out that current annotation schemes are based on psychological theories not primarily focused on text.

\subsection{Inconsistency in Terminology}\label{sec:terminology}
Section \ref{sec:survey_nomenclature} shows the inconsistent terminology used in EA. This lack of specificity can obscure the related but distinct subtasks. Currently, no distinction is made between `emotion classification', for example, and `emotion prediction'. We generally understand \emph{classification} as a subtask of \emph{prediction}, so these terms might be treated as synonymous. We argue that EA should clarify its jargon. 
Given the growing interest in detecting the triggers for emotion in NLP, we suggest using the term \emph{prediction} to demarcate tasks that are interested in emotions \emph{before} they occur, such as forecasting a customer's frustration during a support chat based on contextual cues. One could also make the case that emotion \emph{classification} should refer to classifying a set of emotions, whereas emotion \emph{detection} or \emph{recognition} are searching for a specific emotion for a specific reason. For example, using NLP for suicide prevention might involve detecting \texttt{sadness}, \texttt{desperation}, or \texttt{anger}. However, it might seem odd to say it would involve classifying emotions. 
Mapping EA in NLP using consistent nomenclature will enable NLP practitioners to focus on fine-grained specific tasks since there will be a clear goal. It will also highlight gaps in EA for which new tasks must emerge.

\subsection{Lack of Interdisciplinarity}

Two issues arise out of the general lack of interdisciplinary engagement in EA:
The first issue was already raised by \citet{kusal2022review}, who concluded that it would be helpful to gain a deeper understanding of emotions for the classification process. 
As we showed in Section \ref{ssec:emotion_models}, currently, EA in NLP relies solely on emotion models and frameworks borrowed from psychology. In particular, NLP overwhelmingly relies on Ekman's theory from the 1970s. This theory identifies emotions with bodily signatures, while \citet{barrett2017emotions} has empirically problematized it by stating that emotions do not have such signatures and emphasizes the importance of considering the context for interpreting emotions. Additionally, Ekman's views have been conceptually criticized by philosophers who argue that emotions possess intentionality, being directed towards something, a characteristic not adequately addressed in Ekman's theory.\footnote{For an introductory overview, see \citet{brady2018emotion}. For a more in-depth overview see \citet{sep-emotion}.}


The second issue concerns potential applications across fields. Most papers we reviewed here do not mention the potential applicability of EA to the humanities and social sciences, even though that is a fruitful area of development (e.g., digital humanities). Collaboration between these disciplines can lead to innovative research, the development of specialized tools, and the integration of emotional insights into a broader range of academic studies. This interdisciplinary synergy would be mutually beneficial, enriching the understanding of human emotions and their role in various domains.

\subsection{Demographics and Cultural Implications}

Demographic factors such as age, gender, cultural background, and socioeconomic status can significantly shape individual differences while expressing and experiencing emotions. \citet{bender-friedman-2018-data} call for careful data curation and conscientious reporting of the processes and actors involved in corpus creation. More specifically, they call for reporting who produced and who annotated the data. We note that most papers we reviewed do not include a data statement or report demographics for either the data creators or the annotators. Additionally, when demographic data is collected, it is more commonly associated with annotators rather than data creators, as observed in prior studies (\cite{troiano-etal-2019-crowdsourcing,haider-etal-2020-po,kajiwara-etal-2021-wrime}.   However, it is noteworthy that in most cases, this demographic information is not integrated into the model inputs. Given the subjective nature of emotions and the enormous role annotators and sources can play, this observation highlights a concerning gap in the field. It also emphasizes the need to consider the influence of annotator demographics for a more comprehensive and nuanced approach to EA. Furthermore, recent work \cite{basile-etal-2021-need,plank-2022-problem,davani-etal-2022-dealing} has suggested leveraging human label variation when annotating and creating data. Subjective tasks like EA would benefit from this approach as it allows for developing more inclusive and diverse systems considering individual perspectives and using annotation to explore the range of possible emotions in a task \cite[the descriptive paradigm of][]{rottger-etal-2022-two}. 

Furthermore, uncovering biases and social stereotypes in LLMs requires considering both emotions and demographics. \citet{plaza2024angry} conduct a thorough investigation into gendered emotion attribution across different LLMs, exploring whether emotions are gendered and whether these variations are based on societal stereotypes. Their findings reveal a consistent pattern across all models, indicating gendered emotional attributions influenced by stereotypes.

\section{Discussion}

We can now answer the four questions from the beginning of our study:

\begin{enumerate}
    \item \textbf{How are EA tasks defined in NLP?} The commonly used terms to refer to EA tasks include emotion detection/classification/recognition/prediction being \emph{emotion recognition} the most used. These terms are often used interchangeably in different research papers. Consequently, this interchangeable usage lacks specificity and can potentially obscure the distinct, yet interconnected, subtasks involved.
    \item  \textbf{What are the most prominent emotion frameworks and which emotions are modeled?} The majority of studies primarily rely on Ekman's basic emotions, with Plutchik's model being used to a lesser extent. Consequently, other psychological theories are ignored to some degree. This limits the usefulness of EA models in downstream applications. Recent studies based on cognitive appraisals show future directions to  encode categorical models. Future work should endeavor to diversify the emotions and emotions frameworks represented. 
    \item \textbf{Is the subjectivity of emotions considered in terms of demographics and cultural factors?} No consideration is given to the subjectivity of emotions in relation to demographics and cultural factors. The subjective nature of emotions in EA restricts the diversity and inclusivity of NLP methods. Therefore, it is crucial for future endeavors in dataset creation and NLP system development to take these factors into account.
    \item \textbf{What are the primary NLP applications for EA?} Among the papers reviewed, the majority of them focus on emotion recognition in conversation which involves the analysis of emotions expressed during interactions and generating empathetic responses\footnote{For a thorough discussion on this issue, see \citet{curry2022computer}.}. Additionally, there are applications in areas such as public health, critical event analysis, email customer care, and financial forecasting.
\end{enumerate}

\section{Roadmap for Future Directions}


\paragraph{Demographics for Diversity and Inclusivity.}
Collecting data on the demographics of annotators or data creators, such as age, gender, ethnicity, and cultural background, provides insights into how individual perspectives shape emotional experiences. Integrating demographic information into the NLP model inputs could lead to more context-aware and tailored EA. Research has shown that our perception of emotions is strongly gendered, racialized, and age-dependent, and datasets should capture this diversity. Moreover, while demographics can affect how emotions are perceived, individual differences also impact this~\cite{orlikowski2023ecological}. Avoiding the aggregation of labels and instead focusing on individual demographics promotes a more nuanced and inclusive approach to EA.\footnote{See the Perspectivist Data initiative \url{https://pdai.info/}} 

\paragraph{Tailor Emotion Categories to your Task.} When choosing emotion models or categories, it is vital to not solely depend on the widely repeated theory in the literature. Instead, one should prioritize considering the specific domain and application scenario. In addition, to avoid fixed categorization and in line with standard supervised learning objectives, exploring models that can transfer knowledge across different emotions is desirable. For instance, recent zero or few-shot learning paradigms offer promising alternatives that enable more flexible and adaptable approaches in the field of EA \cite{plaza-del-arco-natural}. 


\paragraph{EA Nomenclature.} 
EA is the broader area that comprises EA tasks in NLP. `Emotion detection', `emotion classification', and `emotion recognition' are often used interchangeably as synonyms. However, it can be beneficial to distinguish between these terms based on the specific task at hand. For example, using NLP for mental disorder detection might involve detecting specific emotions as opposed to classifying the emotions involved. In addition, other terms should not be used as synonyms, given that they contribute to a lack of clarity about the task's aims (e.g., `prediction' and `classification'). 

\paragraph{Interdisciplinarity.} EA is inherently interdisciplinary and borrows from social science and humanities. EA should keep up to date with current theories and models of emotion (e.g., the discrete vs.\ dimensional vs.\ componential models). Furthermore, given that NLP's domain is language, it would stand to benefit from a more pluralistic understanding of emotion engaging with philosophical theories of emotion, such as so-called cognitive theories of emotion 
and perceptual theories of emotion.\footnote{For a survey of theories of emotion, see \citet{brady2018emotion}.}

\section{Conclusion}

EA has grown in NLP since the 2010s, with more research on human emotions in numerous fields. Over time, conferences and publications have published various studies on the area, demonstrating its growing popularity.

We reviewed over 150 ACL anthology papers from 2014 to 2022 in this report. We discussed field nomenclature, emotion theories, demographic and cultural subjectivity, and NLP applications.

We found demographic and cultural gaps, inadequate emotion category match to the downstream goal, no standard systematic nomenclature in EA, and no interdisciplinary research. For each gap, we proposed future directions to develop meaningful linkages, facilitate targeted study, and enable NLP emotion modeling for nuance.

\section*{Limitations}

Our survey primarily focuses on papers included in the ACL anthology. However, we acknowledge the existence of other relevant EA papers in the NLP field that may have been published in journals or other conferences. Furthermore, we are aware that there are papers that may employ varying keywords or terminology when describing their work related to EA.

Finally, while we have conducted a thorough analysis of specific aspects related to EA, there are other facets that we have not covered in this review.

\section*{Ethics Statement}

The paper is related to a survey, therefore, it does not involve ethical issues in the methodology.

\section*{Acknowledgements}

The work of Flor Miriam Plaza-del-Arco, Amanda Cercas Curry and Dirk Hovy has been funded by the European Research Council (ERC) under the European Union’s Horizon 2020 research and innovation program (grant agreement No. 949944, INTEGRATOR). They are members of the MilaNLP group
and the Data and Marketing Insights Unit of the Bocconi Institute for Data Science and Analysis.

\section{Bibliographical References}\label{sec:reference}

\bibliographystyle{lrec-coling2024-natbib}
\bibliography{anthology,custom}

\bibliographystylelanguageresource{lrec-coling2024-natbib}
\bibliographylanguageresource{languageresource}

\end{document}